# Integrated three-dimensional reconstruction using reflectance fields

Maria-Luisa Rosas[1] and Miguel-Octavio Arias[2]

[1,2] Computer Science Department, National Institute of Astrophysics, Optics and Electronics,
Puebla, 72840, Mexico
{mluisa,ariasmo}@ccc.inaoep.mx

**Abstract**
A method to obtain three-dimensional data of real-world objects by integrating their material properties is presented. The material properties are defined by capturing the Reflectance Fields of the real-world objects. It is shown, unlike conventional reconstruction methods, the method is able to use the reflectance information to recover surface depth for objects having a non-Lambertian surface reflectance. It is, for recovering 3D data of objects exhibiting an anisotropic BRDF with an error less than 0.3%.

## 1. Introduction

Many of the current methods in 3D computer vision rely on the assumption that the objects in the scene have Lambertian reflectance surface (such property is related to the materials that reflect the same amount of incident energy illumination uniformly over all the surface). Unfortunately, this assumption is violated for almost all real world objects, leading to incorrect depth estimates [1][2].

In the area of computer graphics, the reflection from surfaces is typically described by high dimensional reflectance functions. However, the formulation of analytical models is not always an easy task. An alternative approach to the specification of the reflectance or optical properties of the surface objects by analytical modeling is the capture of this reflectance information from real-world surfaces. The acquisition is carried out, with a camera or array of cameras to obtain a set of data that describes the transfer of energy between a light field of incoming rays (the illumination) and a light field of outgoing rays (the view). Such set of data is known as the Reflectance Field [3].

This document explores the problem of obtaining the three-dimensional reconstruction of objects exhibiting an anisotropic BRDF (the objects material have the property that their reflection characteristics vary to rotations of the surface about its normal) by using a 4D slice of the 8D reflectance field. The 4D slice of the reflectance field is obtained by a camera-projector pair. Our method exploits the property of reciprocity of the reflectance field to impose the epipolar constraint by considering the camera-projector pair as a stereo system. As an example, we show how our method can be used to recover objects with an anisotropic BRDF of their surface. This procedure avoids the need of an analytical model of the reflectance data.

## 2. Theory

2.1 Reflectance field and Light transport constancy

Debevec et al [3] showed that the exiting light field from the scene under every possible incident field of illumination can be represented as an 8D function called the reflectance field:

$$R(L_i(\psi_i); L_0(\psi_0)) = R(\psi_i; \psi_0)$$

(1)

Here, $L_i(\psi_i)$ represents the incident light field on the scene, and $L_0(\psi_0)$ represents the exiting light field reflected off the scene. In order to work with discrete forms of these functions, the domain $\psi$ of all incoming directions can be parameterized by an array indexed by $i$. The outgoing direction corresponding to an incoming direction is also parameterized by the same index, $i$. Now, consider emitting unit radiance along ray $i$ towards the scene (e.g., using a projector). The resulting light field, which is denoted by vector $\mathbf{t}_i$, captures the full transport of light in response to this impulse illumination. This is called the impulse response or the impulse scatter function [4].
All the impulse responses can be concatenated into a matrix $\mathbf{T}$ which is called the light transport matrix:

$$\mathbf{T} = [\mathbf{t}_1 \mathbf{t}_2 \ldots \mathbf{t}_n]$$

(2)

Since light transport is linear, any outgoing light field represented by a vector $\mathbf{L}_0$ can be described as linear combination of the impulse responses, $\mathbf{t}_i$. Thus, for an incoming illumination described by vector $\mathbf{L}_i$, the outgoing light field can be expressed as:
$$\mathbf{L}_0 = \mathbf{T}\mathbf{L}_i \quad (3)$$

The light transport matrix $\mathbf{T}$, is thus the discrete analog of the reflectance field $R(L_i(\psi_i); L_0(\psi_0))$.

## 2.2 Symmetry of the transport matrix

The idea that the flow of light can be effectively reversed without altering its transport properties was proposed by von Helmholtz in his original treatise in 1856 [5]. He proposed the following reciprocity principle for beams traveling through an optical system (i.e., collections of mirrors, lenses, prisms, etc.):

*Suppose that a beam of light* $\mathbf{A}$ *undergoes any number of reflections or refractions, eventually giving rise (among others) to a beam* $\mathbf{B}$ *whose power is a fraction* $\mathbf{f}$ *of beam* $\mathbf{A}$. *Then on reversing the path of the light, an incident ray* $\mathbf{\acute{B}}$ *will give rise to a beam* $\mathbf{\acute{A}}$ *whose power is the same fraction* $\mathbf{f}$ *of beam* $\mathbf{\acute{B}}$.

In other words, the path of a light beam is always reversible, and furthermore the relative power loss is the same for the propagation in both directions. For the purpose of a reflectance field generation, this reciprocity can be used to derive an equation describing the symmetry of the radiance transfer between incoming and outgoing directions $\psi_i$ and $\psi_0$:
$$R(\psi_i; \psi_0) = R(\psi_0; \psi_i) \quad (4)$$

where $R$ is the reflectance field. For the light transport matrix defined in the last section, this implies that the transport of light between a ray $i$ and a ray $j$ is equal in both directions, i.e.
$$T[i,j] = T[j,i] \Longrightarrow T = T^T \quad (5)$$

Therefore, $T$ is a symmetric matrix (See work in [6]).

## 2.2 BRDF

The Bidirectional Reflectance Distribution Function (BRDF) is a projection of the 8D reflectance field into a lower dimension. From equation 1, the 4D reflectance field can be represented as
$$f_r(L_i(\Omega_1); L_0(\Omega_2)) = f_r(\Omega_1; \Omega_2) \quad (6)$$

where $L_i(\Omega_1)$ represents the incident light field on the scene, and $L_0(\omega_2)$ represents the exiting light field reflected off the scene and $\Omega_1$, $\Omega_2$ are incoming and outgoing directions, e.g., $(\theta_1, \phi_1)$, $(\theta_2, \phi_2)$. In essence, the BRDF describe how bright the differential surface $dA$ of a material appears when it is observed from a certain direction and illuminated from a certain direction.

The reciprocity exposed in the last section, the 4D reflectance field can be written as
$$f_r(\Omega_1; \Omega_2) = f_r(\Omega_2; \Omega_1) \quad (7)$$

Some materials have the property that their reflection characteristics are invariant to rotations of the surface about its normal. Such materials are called isotropic. Materials not having this characteristic are called anisotropic. As equation 2 shows, in order to discretize the equation 7, all incoming and outgoing directions in domain $\Omega$ can be parameterized by an array indexed by $i$. We denote the resulting 4D light field by vector $\hat{t}_i$ and this 4D light field is concatenated as
$$\widehat{\mathbf{T}} = [\mathbf{t}'_1 \mathbf{t}'_2 \ldots \mathbf{t}'_n] \quad (8)$$

For an incoming illumination described by vector $\mathbf{L}'_i$, the outgoing light field can be expressed as
$$\mathbf{L}'_0 = \widehat{\mathbf{T}} \mathbf{L}'_i \quad (9)$$

The matrix $\widehat{\mathbf{T}}$ is the discrete analog of the 4D reflectance field. The reciprocity exposed in the last section implies that the transport of light between a ray $i$ and a ray $j$ is equal in both directions, i.e.
$$\widehat{T}[i,j] = \widehat{T}[j,i] \Longrightarrow \widehat{T} = \widehat{T}^{\widehat{T}} \quad (10)$$

## 3. Depth recovery from the 4D reflectance field

Consider the scene configuration in Fig. 1a. All the scene is illuminated by a projector $\mathbf{L}'_i$. A particular point in the scene $p$ will reflect light to the camera C according to equation 9, the outgoing light field represented by the vector $\mathbf{L}'_0$ is the reflected intensity in the direction of C from the point $p$ with normal vector $\hat{n}$. Let $o_1$ and $o_2$ denote the positions of the projector and camera, respectively. The unit vectors $\Omega_1 = \frac{1}{|o_1 - p|}(o_1 - p)$ and $\Omega_2 = \frac{1}{|o_2 - p|}(o_2 - p)$ denote the directions from $p$ to the projector and camera, respectively. Given this configuration, the image irradiance (see [7]) at the projection of $p$ is

$$e = f_r(\Omega_2, \Omega_1)\frac{\hat{n}\cdot\Omega_2}{|o_2-p|^2} \qquad (11)$$

where $f_r$ is the BRDF (4D function).

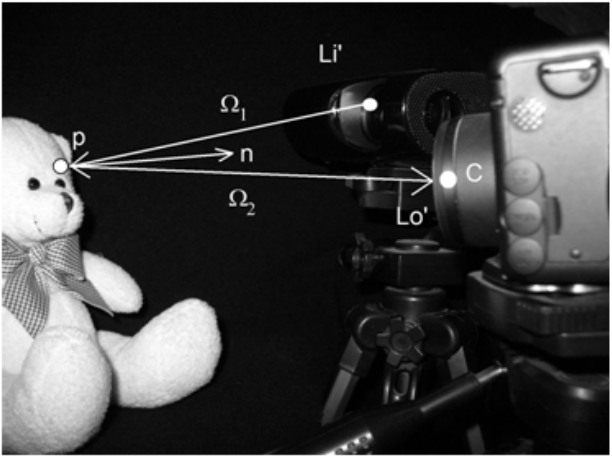

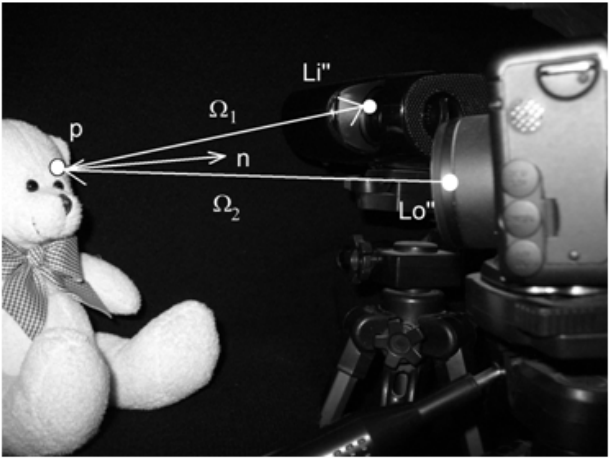

Fig. 1 The scene in (a) is illuminated by a light source $\mathbf{L}'_i$. A particular point in the scene $p$ will reflect light to the camera C. The outgoing light field $\mathbf{L}'_0$ is the reflected intensity in the direction of C from the point $p$. The scene in (b) is "illuminated" by a camera $\mathbf{L}''_0$. A particular point in the scene $p$ will reflect light and it is "captured" by a light source. The outgoing light field $\mathbf{L}''_i$ is the reflected intensity in the direction of light source from the point $p$.

In the above equation is assumed that every ray light $\Omega_2$ from the light source illuminates the scene and the number of rays reflected is just one, it can be considered true for those objects with 4D material properties. Now we add the transport matrix $\widehat{\mathbf{T}}$ to the equation 11, so we have,

$$e = \widehat{\mathbf{T}}(p)\frac{\hat{n}\cdot\Omega_2}{|o_2-p|^2} \qquad (12)$$

where $\widehat{\mathbf{T}}(p)$ is the 4D transport matrix that corresponds to a point $p$ of the scene, $\hat{n}$ can be expressed as $\left(\frac{dz}{dx}, \frac{dz}{dy}, -1\right)$, the ray from the camera can be expressed as $\Omega_2(p) = (\Omega_{2x}, \Omega_{2y}, \Omega_{2z})$

Taking advantage of the symmetry of the transport matrix we can impose the epipolar constraint to provide a solution to equation 12. Consider the scene configuration in Fig. 1b. All the scene is "illuminated" by a camera $\mathbf{L}''_0$. A particular point in the scene $p$ will reflect light and it is "captured" by a light source. Then, we can consider the system configuration as a stereo setup such as, it can be calibrated as a stereo system.

The vector $\Omega_2(p)$ and the denominator $|o_2 - p|^2$ can be determined when calibrating a stereo setup. Imposing the epipolar constraint we can express the normal $\hat{n}$ as $\left(\frac{dz}{dx}, 0, -1\right)$.

The point $p(x, y, z)$ will have projections in the camera and the light source (considered as a second camera) established by calibration parameters of the system. Expressing the depth as $z(x, y)$, we rewrite the equation 12 as

$$\frac{dz}{dx} = \frac{e|o_r-p|^2\widehat{T}(p)+\Omega_2 z}{\widehat{T}(p)\Omega_2 x} \qquad (13)$$

This can be numerically integrated as

$$z(x, y) = \int_{x_0}^{x}\frac{dz}{dx}dx + z(x_0, y) \qquad (14)$$

For each epipolar line $y$, this integral provides the depth across the epipolar line. We can determine for each epipolar line $y$ the $z(x_0, y)$ since the point $p(x, y, z)$ have projections in the camera and we know the corresponding projections to the light source when the 4D transport matrix is captured.

## 3. Test reconstruction

In order to obtain the three-dimensional reconstruction of the object placed in the scene some calibration parameters have to be computed, to do that we use the Dual Photography [6] technique to use the camera-projector assembly as a stereo system for enabling the projector to "capture" images like a camera, thus making the calibration of a projector essentially the same as that of a camera, which is well established. A standard black-and-white checkerboard is used. The flat checkerboard positioned with different poses is imaged by the camera and poses from the point of view of the projector are generated. Once, these poses are obtained the intrinsic and extrinsic parameters of the stereo system using the Matlab toolbox provided by Bouguet [8] are computed.

Fig. 2 shows an example of the checkerboard images captured from the point of view of the camera (a) and synthesized from the point of view of the projector (b).

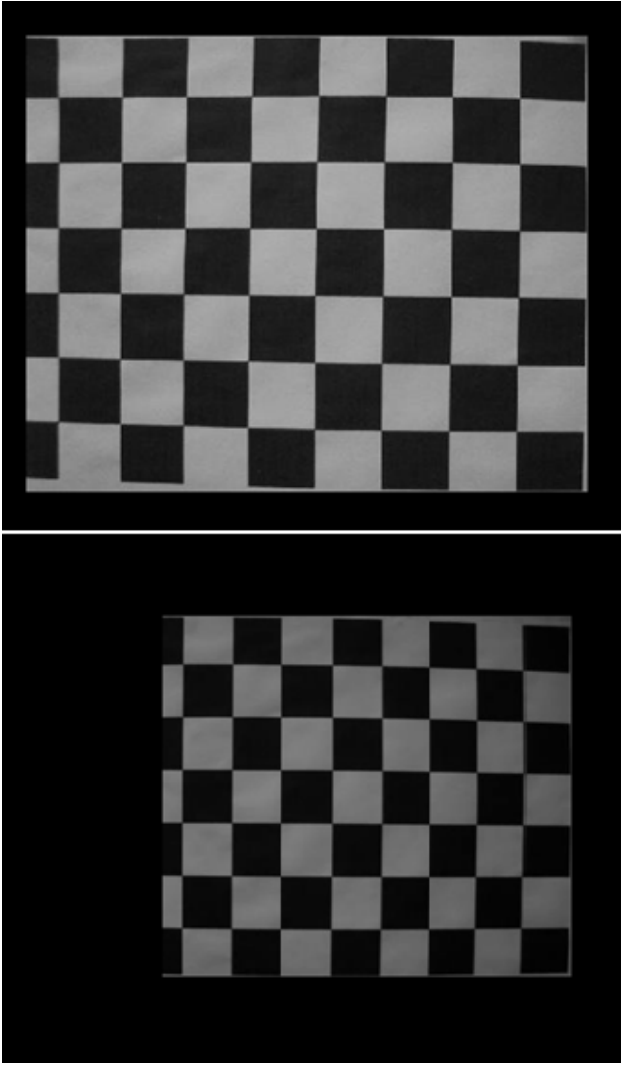

Fig. 2 Images synthesized of the checkerboard from the point of view of the camera (top) and from the point of view of the projector (bottom).

The three-dimensional reconstruction using 4D light field was implemented in Matlab and validated its effectiveness in the following experiment. We obtained the reconstruction of a real object exhibiting a non-Lambertian material which dimensions are known and the RMS error was computed by comparing the reconstructed object and the real object measurements. The Fig. 3 shows the three-dimensional reconstruction of a cube. The RMS error between the cube recovered and the real cube is of 0.3%.

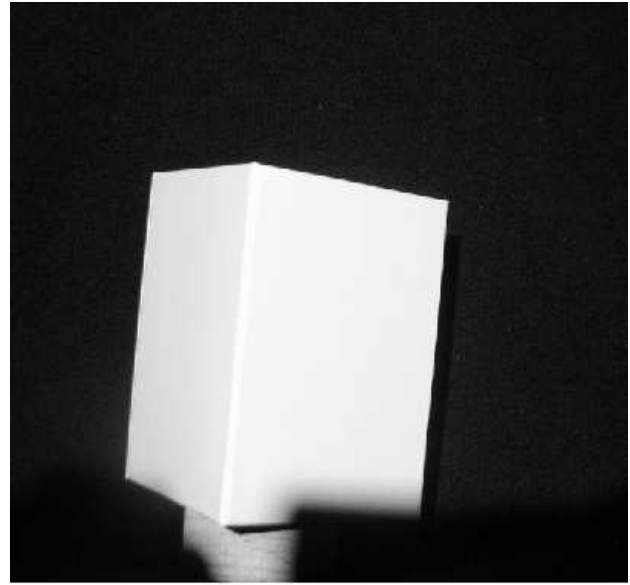

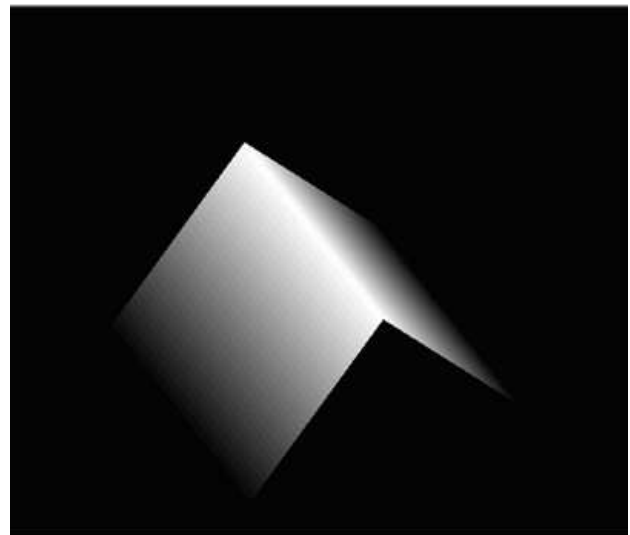

Fig. 3 Three-dimensional reconstruction (b) of a real object exhibiting a non-Lambertian material from its surface (a).

## 4. Conclusions

All methods of three-dimensional reconstruction in computer vision area are influenced by light and the material properties of the objects. The estimation of the material of reflectance properties of the object is important for a correct 3D measurement. In computer graphics, the material properties of such objects materials are measured and they are described as a dimensional reflectance functions (8D function). The theory and experiment have demonstrated the ability of obtaining the three-dimensional reconstruction of objects exhibiting an anisotropic BRDF by integrating a 4D slice of the 8D reflectance field information by using a camera-proyector pair with an error less than 0.3% of the real-world object measurements. Also, this procedure represents the first step to extend the formulation to include 6D and 8D surface properties.

**Acknowledgments**

The authors acknowledge the partial support by Mexican Conacyt.

**First Author** obtained her B.Eng. in Computer Science at the UPAEP (University of Puebla) in Puebla, Mexico, in 2002. She obtained a M.Sc. in Computer Science at the INAOE (National Institute of Astrophysics, Optics and Electronics, Puebla, Mexico) in 2004. For two years (2006-2008), she worked in Prefixa Vision Systems (Puebla, Mexico) where she developed a 3D Camera. Since 2008 she is a Ph.D student in the Computer Science department at the INAOE. She is an inventor of the patent: Method and apparatus for rapid three-dimensional restoration. Her current research interests are computer vision, computer graphics, FPGA and CUDA architectures, robotics and genetic algorithms.

**Second Author** obtained his B.Eng. in Communications and Electronics at the FIMEE (University of Guanajuato) in Salamanca, Gto. in 1990. He also obtained a M.Eng. in Instrumentation and Digital Systems at the FIMEE two years later. In 1997, he finished his Ph.D. degree at the Computer Vision and Systems Laboratory of Université Laval (Quebec city, Canada). He was a professor-researcher at the Computer and Systems Laboratory at Laval University where he worked on the development of a Smart Vision Camera. Since 1998 he is with the Computer Science department of INAOE (National Institute of Astrophysics, Optics and Electronics, Puebla, Mexico) where he continues his research on FPGA architectures for computer vision.